\documentclass[10pt,twocolumn,letterpaper]{article}

\usepackage{wacv}
\usepackage{microtype}
\usepackage{graphicx}
\usepackage{subfigure}
\usepackage{epsfig}
\usepackage{booktabs} 
\usepackage[breaklinks=true,bookmarks=false]{hyperref}
\usepackage{footnote}
\usepackage{amsmath}
\usepackage{makecell}

\wacvfinalcopy

\def\wacvPaperID{399} 

\begin{document}

\title{Receptive Field Size Optimization with Continuous Time Pooling}

\author{
Dóra Babicz, Soma Kontár, Márk Pető, András Fülöp, Gergely Szabó, András Horváth \\
Peter Pazmany Catholic University - Faculty of Information Technology and Bionics\\
Práter u. 50/A, 1083, Budapest, Hungary\\
{\tt\small horvath.andras@itk.ppke.hu}\\
}

\maketitle

\begin{abstract}

The pooling operation is a cornerstone element of convolutional neural networks. These elements generate receptive fields for neurons, in which local perturbations should have minimal effect on the output activations, increasing robustness and invariance of the network.
In this paper we will present an altered version of the most commonly applied method, maximum pooling, where pooling in theory is substituted by a continuous time differential equation, which generates a location sensitive pooling operation, more similar to biological receptive fields.
We will present how this continuous method can be approximated numerically using discrete operations which fit ideally on a GPU. In our approach the kernel size is substituted by diffusion strength which is a continuous valued parameter, this way it can be optimized by gradient descent algorithms.
We will evaluate the effect of continuous pooling on accuracy and computational need using commonly applied network architectures and datasets.


\end{abstract}

\section{Introduction}
The revolution of computer vision was mainly driven by convolutional neural networks (CNNs) in the past decade. These architectures contain an alternating series of convolution and pooling operations. Since then other variants have also appeared, like residual networks \cite{he2016deep} and all convolution networks \cite{springenberg2014striving}, but these two operations can still be found in the applied architectures.
In the paper of LeCun, Bengio, and Hinton \cite{lecun2015deep} originally the alteration of convolution and a pooling layers was motivated by the order of the simple and complex cells in the cortex. The majority of currently applied methods are built upon these two elements.

Convolutions are responsible to detect certain patterns and their weights are trainable by the network, meanwhile pooling operations are parameterless and are responsible for subsampling the activations in the receptive field of a neuron, decreasing the dimensions of  the data and increasing invariance by neglecting minor perturbations. The most commonly applied version is maximum pooling which selects only the largest response and neglects all other values.
These two functionalities: dimension reduction and  disregarding unimportant changes can be separated from each other and one can easily see, that the second element is way more important and affects the accuracy and the robustness of a network significantly. 
Subsampling can be implemented by simply indexing out the necessary elements or using any kind of windowed operation with larger strides than one.
The optimal size of the receptive field is a key element in all networks regarding invariance and it is difficult to be optimized, since kernel sizes are discrete values, which are hyper-parameters and can only be tested by training a new network with the modified values.

In biology certain neurons can receive inputs from different number of neighbours and the size of their receptive field may also vary \cite{wang2010visual}.
It can also be observed that unlike in the case of maximum pooling, the response of a neuron is not completely independent from the position of the activation inside the receptive field.
In many cases, especially in regions that have visual information processing tasks, the effect of these spatial formations shows high similarity to Gaussian, Difference of Gaussians (DOG) or Laplacian of Gaussian (LOG) functions as it is stated in \cite{lindeberg2013computational}.
This demonstrates, that at least in certain parts of the brain or the retina, the neuronal behavior of receptive fields is very much continuous in space and spatial location does matter \cite{lindeberg2013computational}. 
This motivated us to implement and investigate a version of maximum pooling, where the largest activation is selected, but apart from its value the location of the activation is also taken into account.
This can be considered as weighting the activation with a Gaussian distribution of its relative position.
Those activations which are in the central region of the kernel produce the highest activation and a small translation will result in a small decrease in the response.
We will also show that applying a continuous distribution instead of discrete sized window for pooling, one can optimize the size of the pooling kernel as a continuous variable and this way it can be tuned by gradient based optimization methods, rather than optimizing them as hyper-parameters.
We implement our method, called continuous time pooling using a non-linear diffusion operator where the propagation time of the diffusion and the diffusion coefficient, which we refer as pooling strength, will determine the size and shape of the pooling kernel.

Our paper is organized as follows.
In Section \ref{SecPooling} we will briefly describe the most commonly used pooling operators and we will also introduce our approach: Continuous time pooling.
In Section \ref{SecWeakness} we will introduce a simulated dataset which reveals two different weaknesses of pooling operations and will also present how our approach can solve these tasks.
In Section \ref{SecExperiments} we will validate our method using commonly applied databases. In Section \ref{REcFieldQuant} we will demonstrate how the extra computation arising from our method can be avoided during inference and in Section \ref{SecConclusion} we will conclude our results.

\vspace{-0.2cm}
\section{Pooling Operations}\label{SecPooling}

CNNs are built up by three main operations: Convolution, Non-linearity and Pooling.
We have to note that there are many more commonly applied elements and layers like batch normalization \cite{ioffe2015batch}, dropout \cite{srivastava2014dropout} and others which can be used to increase the generalization capabilities and robustness of a system or other structural and architectural variants like residual networks \cite{he2016deep} or all convolutional networks \cite{springenberg2014striving}, but these three elements are found in almost every commonly used architecture.
Common networks are built as an acyclic graph of these operations. The topology of this graph along with the operations at the nodes determines the network's complexity and by this the set of those functions which the network can approximate. 

Average pooling was the first commonly applied subsampling operation, where a feature map was down-sampled according to a sampling window and each region was substituted by its average value:
\begin{equation}
   P_{avg}(I_{i,j})=\frac{1}{N} \sum_{k,l \in R_{i,j}} I_{k,l}
\end{equation}
where $P_{avg}$ is the average pooling operator. $I$ is the input feature map, $R$ is a two-dimensional region which is selected for pooling and $N$ is the number of elements in the region. 
The notation describes two-dimensional pooling and feature maps, because the operation is most commonly applied on images, but can also be used with one- or higher-dimensional data as well.
This notation focuses only on the pooled region and does not deal with the stride of the convolution operation, which can be used to set the overlapping area between pooling regions, but we consider this as a hyper-parameter of the network architecture and not an inherent part of the operator.

\begin{figure}[h]
    \centering
    \subfigure{
    \includegraphics[width=3.3in]{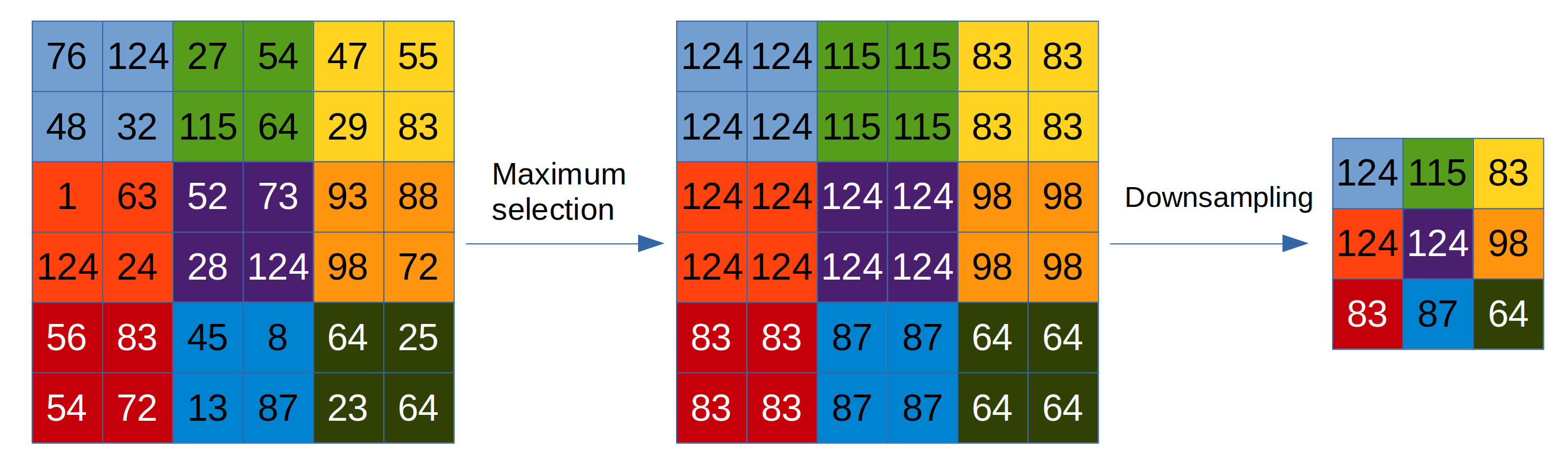}
    }\\
    \subfigure{
    \includegraphics[width=3.3in]{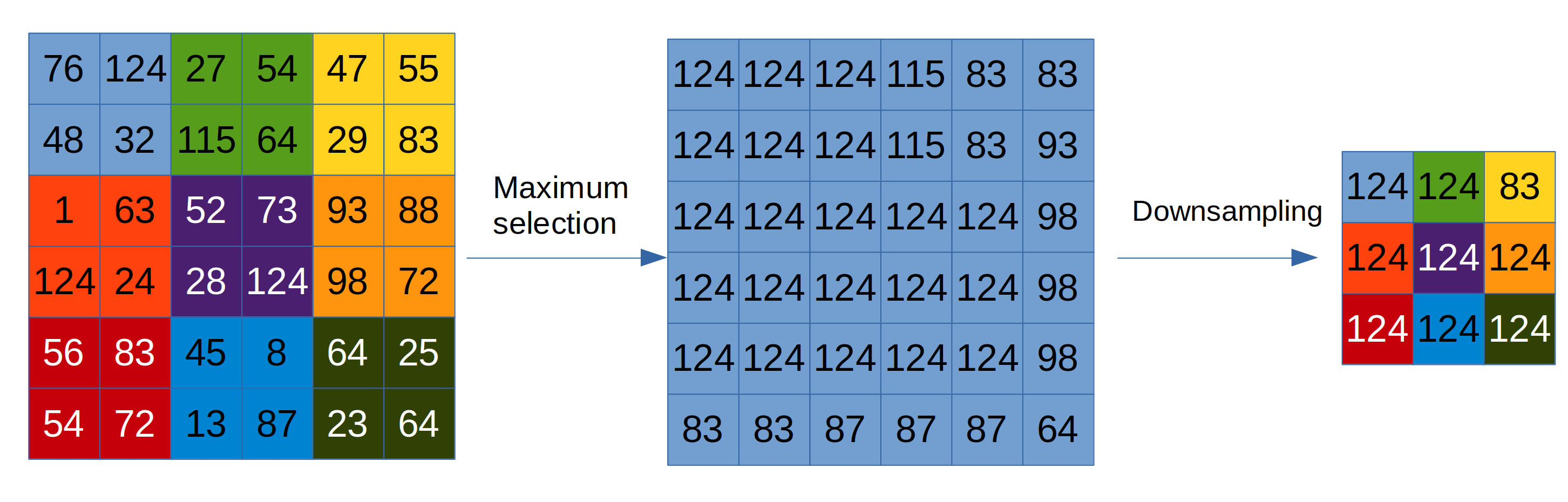}
    }
    \caption{Example of maximum pooling, dividing the operation into two steps. The first step selects the maximum value in each window and the second step selects a single element (for example top left) from each of these windows. In this example the strides of the pooling operation ($2\times2$) is the same as the window sizes ($2\times2$), creating no overlap between the windows. As it can be seen this operation can introduce quantization error, when pixels at the boundary of the window can move to a neighbouring window caused by a small shift (one pixel). If the two operations are separated the window of maximum selection end the stride used for downsampling can be different. The lower example depicts maximum selection with $3\times3$ windows with a stride of one. (Regions are not coloured in this case since it would be confusing because of overlapping windows) and downsampling is implemented by selecting the top left pixel in every $2\times2$ window. In this case robustness towards small changes and the receptive field of a neuron is determined by the first operation. As one can see from this example a small change in the hyperparameters of the pooling method can result a drastically different activation map.}
    \label{fig:maxpool}
\vspace{-0.5cm}
\end{figure}

Average pooling considers the whole input region and all values are used to create the response of the pooling layer which is beneficial for the selection of general features. On the other hand average pooling is a linear operator and can be considered a simple and special case of a convolution kernel with uniformly distributed values and two subsequent convolutions can be replaced by one larger convolution kernel. As novel applications developed, average pooling was substituted in almost every case by maximum pooling. Where the maximum activations are selected from each region:
\begin{equation}
    P_{max}(I_{i,j})= \max_{k,l \in R_{i,j} }(I_{k,l})
\end{equation}
The operation of maximum pooling is depicted on Fig. \ref{fig:maxpool}.

Maximum pooling performs well in practice, adds extra non-linearity to the network and can be efficiently calculated (only an index has to be stored to propagate the error back).
Maximum pooling also has disadvantages. It results in a really sparse update in the network. Only the neuron resulting the maximum element in the kernel will be responsible for the update of the variables during backpropagation and the activation of other neurons does not matter at all. There is a tendency in certain networks like variational autoencoders \cite{kingma2013auto}, generative adversarial networks (GANs)\cite{goodfellow2014generative}, or networks used for one shot learning \cite{fei2006one} to avoid pooling because it grasps only certain elements of the input data  which results in features which add an aliasing effect to generated images.
Modified pooling methods has also appeared in segmentation problems like region of interest pooling in \cite{ren2015faster} or \cite{he2017mask} which helps in the more accurate localization of regions of interest, but does not help in the selection of features inside the proposed regions, where usually maximum pooling is used.
Another approach is introduced in \cite{murray2014generalized} which uses patch similarity over batches to generate a combination of weighted and maximum pooling. \cite{lee2016generalizing} also proposes the application of heterogeneous pooling methods inside the network (e.g.: average pooling in certain layers and max pooling in other) but the selection of pooling methods for each layer is a hyper-parameter of the network and is difficult to be optimized in practice.
The most recent improvements in pooling introduced methods in which the local importance and structure of feature maps should also be preserved. The conversation of local information is implemented in a handcrafted and statistical way in detail preserving pooling \cite{saeedan2018detail} and is further improved by applying  trainable, local attention maps in Local Importance-based Pooling (LIP) in \cite{gao2019lip}.
Other approaches such as Wavelet Pooling \cite{williams2018wavelet} and Wavelet Integrated CNNs \cite{li2020wavelet} incorporate Discrete Wavelet Transform into the pooling operation to preserve the global structure of activation maps better and provide increased robustness to noise.

The problem from a different aspect was also recently demonstrated in \cite{zhang2019making}, where the authors expose that convolutional networks are not shift equivariant to small shifts because of the pooling operations. In \cite{zhang2019making} anti-aliasing with a fixed kernel is proposed as a possible solution, which could mitigate this problem. In this paper we will propose the solution which approximates maximum selection as a discretized version of a continuous process and by this both the width of the pooling window and the anti-aliasing kernel could be learned by the network.

\subsection{Continuous time pooling}

While maximum pooling yields invariance to small perturbations in the receptive field of a neuron, one could observe that biological receptive fields in certain cases are sensitive to the location of the activation and meanwhile shifting an activation in and out of on/off regions can cause drastic differences, shifts inside the receptive fields can also cause minor changes in activations \cite{lindeberg2013computational}.

\begin{figure}[ht]
\vskip 0.2in
 \centering
    \subfigure{
    \includegraphics[width=1.2in]{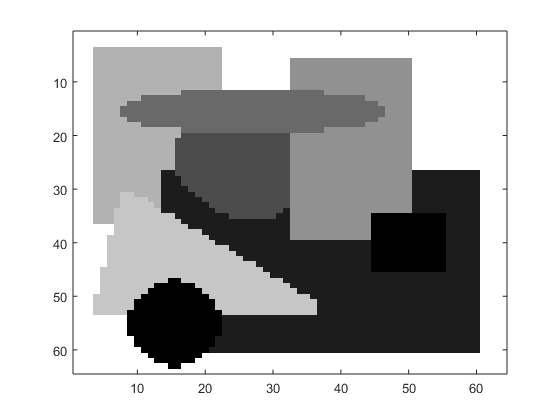}
    \includegraphics[width=1.2in]{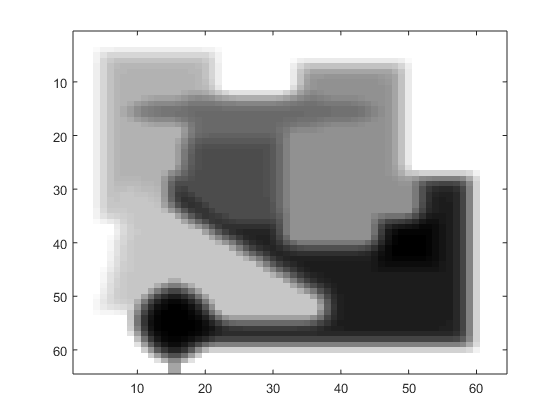}
    }\\
    \subfigure{
    \includegraphics[width=1.2in]{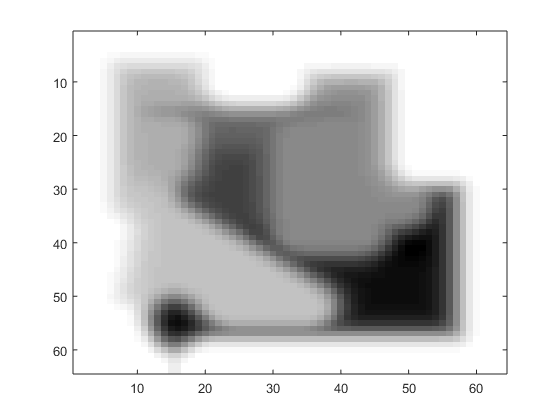}
    \includegraphics[width=1.2in]{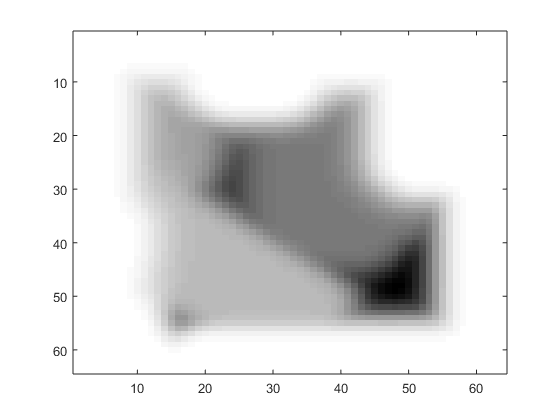}
    }
    \caption{This figure depicts the results of the non-linear diffusion with different execution times. The first image was taken a $t=0$, on the second, third and fourth images the diffusion was executed for $0.2$, $0.5$ and $1$ time units accordingly. As it can be seen in the figure each pixel is slowly and continuously overwritten by the largest intensity in its neighbourhood. These images were generated using a continuous time architecture: a Cellular Neural Network.}
\label{fig:diffusion}
\end{figure}

Continuous time pooling can be considered as the continuous time counterpart of maximum pooling. It can be represented by a Gaussian window, with learnable width implemented by a diffusion operator and can be defined by the following differential equation: 
\begin{equation}\label{diff_eq}
\frac{ \partial I_{i,j}}{ \partial {t}} = S \sum_{k,l \in R_{i,j}}max(I_{k,l}(t) - I_{i,j}(t),0)
\end{equation}
where $I_{i,j}$ is the pooled value of neuron $i,j$ and the same operation is executed simultaneously for every neuron in the array. $I$ is initialized by the image on which the pooling operation is executed. $S$ is a parameter, the time constant of the diffusion. The maximum operation inside the summation is a ReLU operation, which results that the derivative can only be positive, and by this $I_{i,j}$ will converge to the largest intensity in $R_{i,j}$.
In this case the width of the pooling window is determined by the execution time of the diffusion operator, meanwhile the shape of the pooling kernel will be determined by $S$. The result of this non-linear diffusion is depicted in Fig. \ref{fig:diffusion}.

This operation can be efficiently implemented on continuous time neural networks, like Cellular Neural Networks \cite{horvath2017cellular}.
Although these architectures provide an interesting theoretical background for neural networks, they are not commonly applied, since discrete time computers are easier to manufacture and are more robust to noise in practice.
Because of this, we also present the discretized approximation of this operation, which can be derived by approximating Eq. \ref{diff_eq} with the Euler method:
\begin{equation}\label{discretized_same}
I_{i,j}(t+1) = I_{i,j}(t) + hS \sum_{k,l \in R_{i,j}} \max(I_{k,l}(t) - I_{i,j}(t),0)
\end{equation}
Where $h$ is the time step of the Euler method. $t$ goes from $0$ to $N$, $I(0)$ is the map of the input activations and $I(N)$ is the output map of the pooling operation. We have to emphasize that this operation does not implement downsampling, it only selects intensities in the input. Downsampling can be implemented by indexing out the appropriate elements or using any windowed operation (such as maximum or average pooling) with a stride larger than one.

This operation can also be approximated by:
\begin{equation}\label{discretized_appr}
I_{i,j}(t+1) = I_{i,j}(t) + hS\max_{k,l \in R_{i,j}}( \max(I_{k,l}(t) - I_{i,j}(t),0))
\end{equation}
Where instead of summing all the positive differences between the pixels one uses only the largest positive difference. One can easily see that $I_{i,j}$ converges to the same value in Eq. \ref{discretized_same} and Eq. \ref{discretized_appr} which is the maximum intensity pixel in both cases.

Eq. \ref{discretized_same} and Eq. \ref{discretized_appr} are discretized versions of the diffusion equation defined in Eq. \ref{diff_eq}  and as it can be seen, the maximum intensity propagates through the input data in a diffuse way and in continuous time (Eq. \ref{diff_eq}) the distance of propagation is defined by the execution time (the length of the diffusion). However, in discrete domain the size of the pooling window is determined by the number of iterations ($N$) multiplied by the step size of the approximation ($h$), meanwhile window shape is determined by $S$. By setting a constant number of iterations $N$, one can tune the execution time by optimizing the multiplication parameters time step ($h$) and Pooling strength ($S$). Since this is a numerical approximation, $h$ has to be small enough to provide an appropriate solution and any smaller $h$ value will also result an accurate approximation. Setting $h$ two orders of magnitude smaller and increasing $N$ keeping $Nh$ the same value, the approximation will result the same output, meanwhile ensuring that $S$ can increase two orders of magnitude and still result accurate approximation.

Since execution time depends on the multiplication of these two values one can combine them and introduce: $P_S=h S$, which we will refer to as pooling strength.
$P_S$ could have different values for every layer, feature map or even for every neuron. Since convolutional networks should be shift invariant, we opted out from using different $P_S$ for each neuron, since this way the receptive field of a neuron could depend on its location. We have set different $P_S$ parameters optimized individually for each depth dimension after convolutions, but the authors plan to investigate the effect of a layer and neuron dependent implementations as well. We have also used different $P_S$ parameters at each iteration of the Euler method: $P_{S_t}$. This way vanishing and exploding gradients are avoided since the high number of iterations will not be derived back to the same parameter.

\begin{figure}[ht]
\vskip 0.2in
 \centering
    \subfigure{
    \includegraphics[width=1.2in]{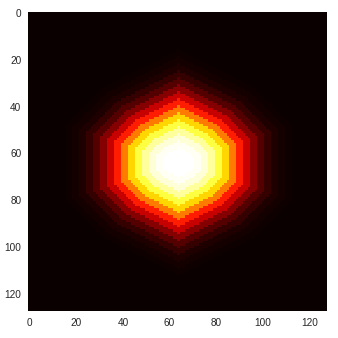}
    }
    \subfigure{
    \includegraphics[width=1.2in]{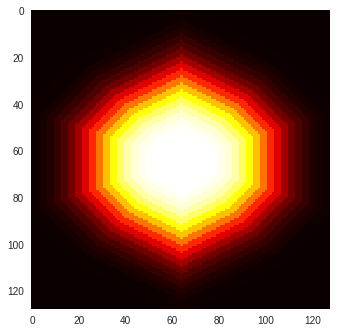}
    }\\
    \subfigure{
    \includegraphics[width=1.2in]{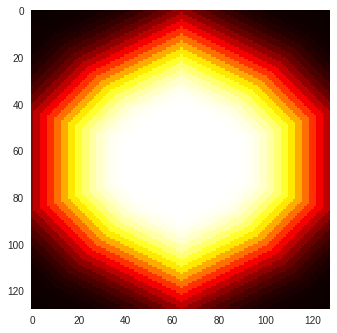}
    }
    \subfigure{
    \includegraphics[width=1.2in]{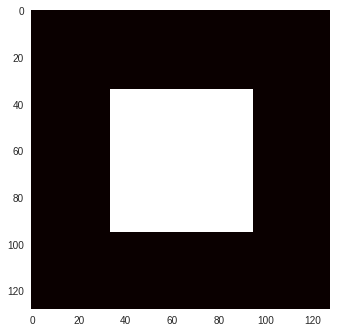}
    }
    \caption{The first three images depict three different continuous time poolings with different pooling strength parameters on a completely black image of size 128x128, containing only a single white pixel in the middle.
As it can be seen both the width and the shape of the pooling window can be changed by modifying the continuous parameter $P_S$. The first three images from the left depict activations of a Dirac delta input with $P_S$ equals 0.0008, 0.0012, 0.0018 (from left to right).The number of iterations for all continuous pooling operations was set to 10000 steps.
The last image depicts the standard implementation of maximum pooling with a window size of 30 pixels and stride of 1.}
\label{fig:continuous_pooling_plot}
\vskip -0.1in
\end{figure}

The effect of $P_S$ on the width and shape of the pooling window can be seen on Fig. \ref{fig:continuous_pooling_plot} along with a comparison to the commonly applied maximum pooling operator.

As it can be seen the window size of maximum pooling is a fixed, discrete value and can only be optimized as a hyperparameter, meanwhile $P_S$, which corresponds to the learnable width of the continuous pooling layer can take arbitrary values and the window width can change continuously, which gives an opportunity for optimization using gradient based algorithms. 

\subsection{Gradient calculation for Continuous time pooling}

Our approximation of the diffusion like continuous time pooling operation (Eq. \ref{discretized_appr}) is built up from simple operations: maximum pooling, addition and multiplication. Its derivative can also be calculated simply and these operations can be found in all machine learning frameworks. 

We would like to emphasize the fact that the derivative is determined not only by intensities, but also by the relative position of the intensities with respect to the investigated pixel. In the first iteration the largest of those pixels are considered, which are direct neighbours of neuron $i,j$ in the next iteration the neighbours of the direct neighbours are also involved and so on. During these calculation the closer the maximum intensity is to neuron $i,j$ the more iterations will it be involved.
Continuous pooling enables the network to have neurons with larger receptive fields, which helps extract more general features thus improving the generalization ability of the network. As stated earlier, in the proposed pooling operation the maximum activation is selected from the current pooling window ($R_{i, j}$), but the result is weighted by the relative distance of the activation. 
Continuous time pooling, unlike maximum pooling, can be implemented in the continuous time domain, and is defined by a differential equation which is in our case approximated by the Euler method. This enables us to employ iterative, gradient-based optimization methods to tune the size of the pooling kernel.
The weighting and the window size is determined by $P_S$ which can be optimized along with the other parameters of the network.
However, to enable the continuous pooling operation to function as a drop-in replacement for maximum or average pooling, we have added an additional element to downscale the data. This additional operation can either be implemented as maximum pooling, average pooling or strided convolution. We have found that the choice of this aggregation operation does not have a large impact on the accuracy of the network which is more determined by the size of the receptive field of the neuron.

\vspace{-0.2cm}
\section{Caveats of Maximum pooling}\label{SecWeakness}

To demonstrate the limitations of maximum pooling we have generated a simple setup where completely black images with dimensions of $32 \times 32$ were generated containing only two non-zero pixels with intensity of one. The task was a simple regression, where we have tried to train a neural network to estimate the squared distance between the two non-zero pixels.

The architecture chosen from this test was the LeNet-5 \cite{lecun1998gradient} containing a single neuron in the output layer representing the distance between the two non-zero points. We have trained two variants of the network with maximum and continuous time pooling, to minimize the mean squared error between the real distances and the estimation of the networks using batches of 32 with AdamOptimizer \cite{kingma2014adam} with an initial learning rate of $10^{-4}$. For continuous pooling we have used ten iterations ($N=10$) and all $P_{S_t}$ parameters were initialized with values of $0.1$.
Our database did not have previously generated train and test sets, but all images were randomly generated during the calculations both for network training and evaluation. During training 64000 different images were presented to the network and after each training step accuracy was evaluated using 1000 independent test samples, which were not used for parameter optimization.

\begin{figure}[ht]
\vskip 0.2in
 \centering
    \subfigure{
    \includegraphics[width=2.7in]{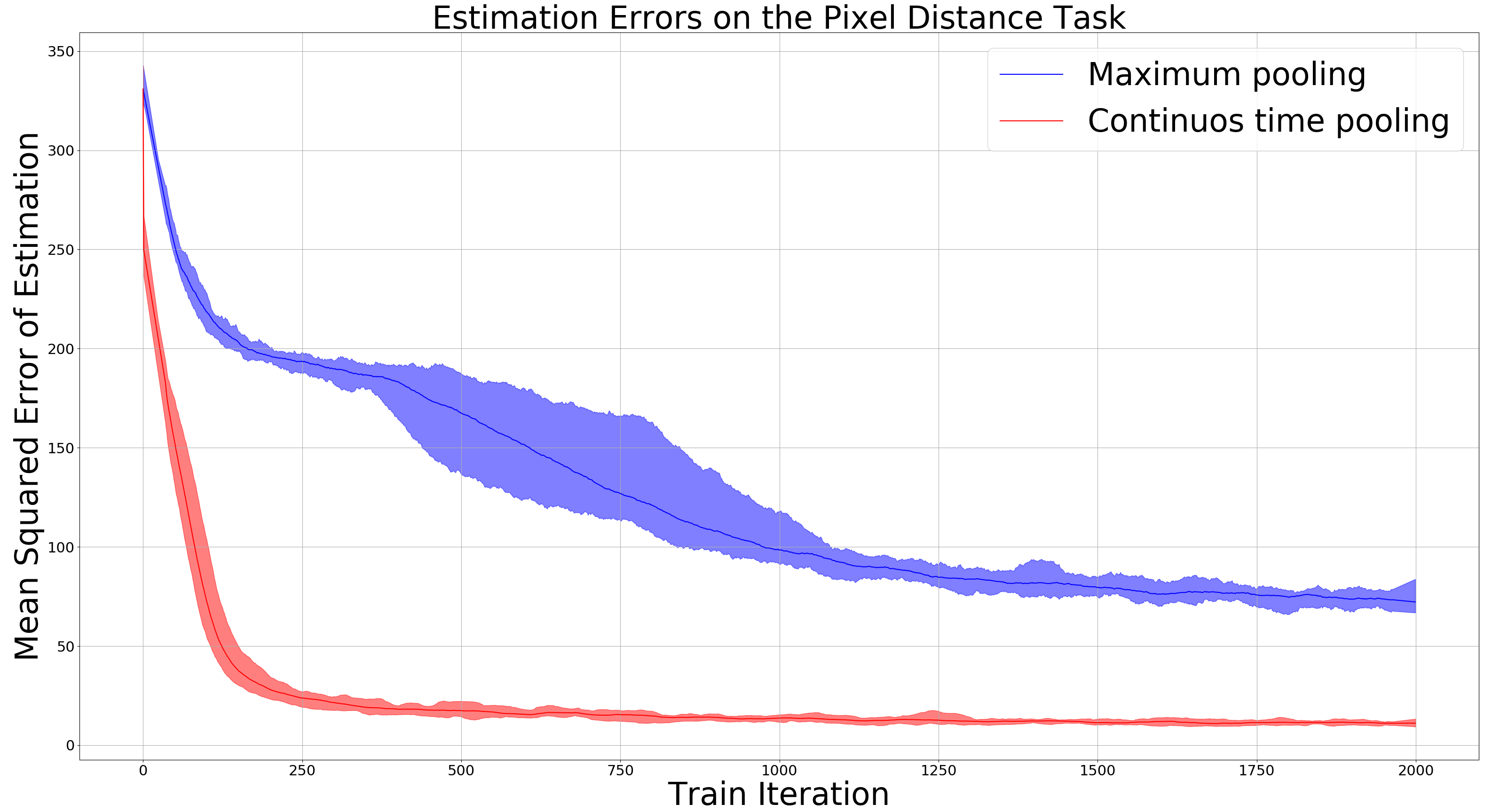}
    }
    \caption{This figure depicts maximum pooling and continuous time pooling using the LeNet-5 network on the pixel distance estimation task. The blue plot depicts test errors for the regular LeNet-5 architecture, meanwhile the red curve shows mean squared errors for continuous pooling. The two lines are averaged results over 20 independent trainings and the shaded intervals depict minimal and maximal values in the trainings.}
\label{fig_pixel_dist}
\vskip -0.2in
\end{figure}

During our first test the non-zero pixels could occupy any positions on the input images.
The result on this dataset can be seen on Fig. \ref{fig_pixel_dist}.

In this setup the LeNet-5 architecture achieved a mean squared error of 70, meanwhile this error decreased to 10 in case of continuous time pooling. 
One could argue that this comparison is not fair, since neurons in the last convolutional layer of LeNet-5 have a receptive field of seven, which is not enough to cover all possible distances. We also completely support this argument, but by this we wanted to demonstrate how important receptive fields are. Using continuous time pooling the network can automatically increase receptive fields and optimize them to the task and the training data.
The distribution of the learned $P_{S_t}$ parameters were ranging between $0.564$ and $2.326$ for the first layer and $0.626$ and $3.775$ for the second layer, which reveals that the optimization has increased the sizes of the receptive fields in the network.

To compensate the bias of the receptive fields, we have created a new task, where the maximal squared distance between the two non-zero pixels had to be less than $49$. In this case it was ensured that both pixels are in the receptive fields of the neurons of the last fully connected layer.
By this we have eliminated the factor that inaccuracy could come from the bad tuning of hyperparameters and the important pixels could not be in the receptive field of the same neuron.

\begin{figure}[ht]
\vskip 0.2in
 \centering
    \subfigure{
    \includegraphics[width=2.7in]{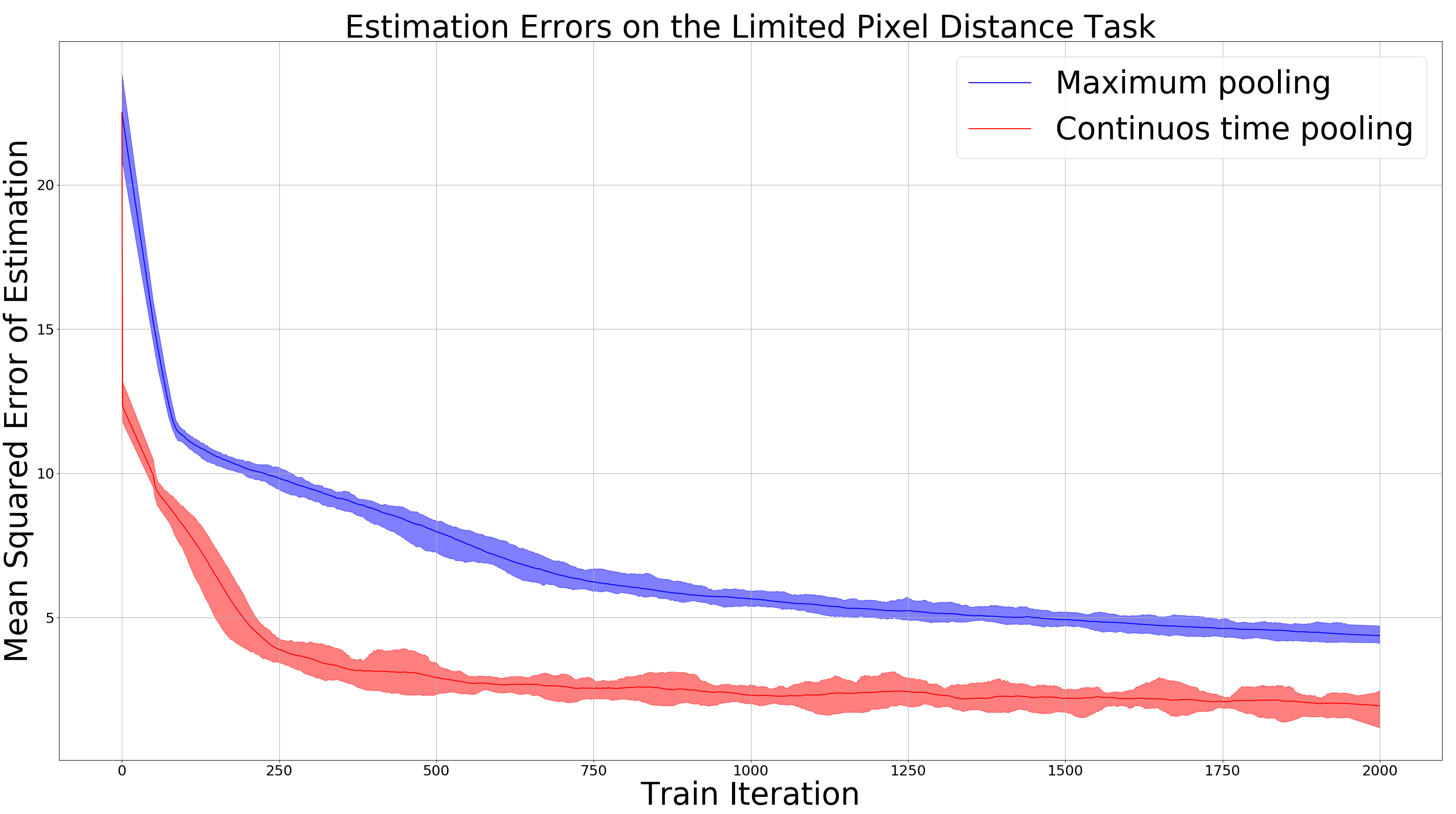}
    }
    \caption{This figure depicts maximum pooling and continuous time pooling using the LeNet-5 network on the limited pixel distance estimation task. The blue plot depicts test errors for the regular LeNet-5 architecture, meanwhile the red curve shows mean squared errors for continuous pooling. The two lines are averaged results over 20 independent training and the shaded intervals depict minimal and maximal values in the trainings.}
\label{fig_limited_pixel_dist_2}
\vskip -0.2in
\end{figure}

Even in this case continuous time poling significantly outperformed its regular counterpart as it can be seen on Fig. \ref{fig_limited_pixel_dist_2}.
On this dataset our method has reached average mean squared error of $2.1$, meanwhile maximum pooling has achieved an error of $4.3$, which is more than double of our error.
This difference demonstrates a serious flaw of maximum pooling, that the location of the detected intensity is completely neglected and by using maximum pooling, the network has to learn different features for every distance.
\vspace{-0.2cm}

This task shows that approximation of distances is difficult for convolutional networks, since a different feature map, convolutional kernel has to be used for each distance.
If an activation is inside the kernel its position does not matter at all, it is difficult to identify the optimal kernel size for maximum pooling as a hyperparameter (in case of LeNet-5 it is set to 2x2) which in case of continuous pooling might be optimized using gradient based methods. 

\vspace{-0.2cm}
\section{Experiments}\label{SecExperiments}

\subsection{MNIST}\label{MNISTtest}

The experiments on the simulated data set in the previous section revealed weaknesses of maximum pooling in convolutional neural networks. Based on our findings we have tried to demonstrate continuous time pooling on the MNIST dataset \cite{lecun1998mnist} using the LeNet-5 \cite{lecun1998gradient} architecture. We have trained the network with exactly the same training parameters as in the previous section, but used the outputs for digit classification (the last layer was reverted back to the original ten neurons of LeNet-5). 
We will only list the most important hyperparameters of our training algorithms, but we would like to emphasize that our training scripts and codes can be find on Github: \href{https://github.com/horvathan/ContinuousPooling}{ \url{https://github.com/horvathan/ContinuousPooling}}

The train and test accuracies averaged over ten independent trainings compared to LIP and Wavelet Pooling can be seen in Table \ref{TableImgClassificationResults}.
As it can be seen the network reaches $95.30\%$ accuracy with maximum pooling. which is the same cited by Lecun in the original paper \cite{lecun1998gradient}. LIP performs poorly on this dataset ($96.65\%$), since the local structure of extremely high and low intensities of the MNIST dataset is already well preserved my maximum pooling. Our variant with continuous pooling achieved $98.87\%$ with the same network architecture and exactly the same training parameters, meanwhile Wavelet Pooling resulted the highest accuracy: $99.01\%$.

\subsection{CIFAR-10}
We have also investigated the AlexNet \cite{krizhevsky2012imagenet} architecture on CIFAR-10 \cite{krizhevsky2014cifar} and compared the same pooling operations as in Sec. \ref{MNISTtest} to our continuous pooling operation. All training parameters, optimizer (gradient descent with momentum) and batch size (128) were the same for every pooling operation.

The original $32\times32$ images of CIFAR-10 were rescaled to $227\times227$ to ensure the appropriate input dimensions for AlexNet.
We have trained our networks for 100 epochs and measured their classification accuracy on the test set. The average accuracies of ten independent runs can be seen in figure Table \ref{TableImgClassificationResults}.
On this dataset continuous pooling heavily outperformed all other pooling variants with an increase of $3.56\%$ in classification accuracy.
We suspect that this is partially caused by the extreme upscaling of the low-resolution images, which resulted the appearance of image features in various distances from each other, but this requires further investigation.


\subsection{ImageNet}

We have selected the ImageNet 2012 \cite{deng2012imagenet} dataset for further investigation and examined three different network types:  the VGG-16 \cite{simonyan2014very}, DenseNets \cite{huang2017densely} and Residual Networks \cite{he2016deep} (ResNet-50 and ResNet-101 architectures) and the previously mentioned four pooling methods.

Training was implemented with batches of 256 following the training method suggested in \cite{goyal2017accurate} using SGD optimizer with Nesterov momentum and weight decay of $10^{-4}$ for four million training iterations.
The top-1 accuracies on the validation set of ImageNet 2012 are reported in Table \ref{TableImgClassificationResults}.

As it can be seen from the results continuous pooling resulted the highest top-1 accuracy in case of the VGG-16, DenseNet-121 and ResNet-50 architectures and a comparable performance using the ResNet-101 architecture, which resulted the highest top-1 accuracy using LIP.

Here we also have to report the additional computational need of maximum pooling. Since our operation is implemented as an iterative approximation of a differential equation it requires multiple (in this case $10$) steps which are equal to maximum pooling in computational complexity. This way our method requires ten times the computational need of maximum pooling, which is a large increase in the number of operations and as the number of steps increases in the approximation, our operations require even more computation. Fortunately the pooling operation is not the bottleneck operation in commonly applied convolutional networks. We have measured the wall time of an average training iteration using VGG-16 on an NVIDIA RTX 2080 TI and using maximum pooling one iteration required $149ms$ in average, meanwhile changing all pooling operations to their continuous counterparts and approximating them with 10 iterations has increased it to $280ms$ in average, which results an $87\%$ increase in training time. This difference was the most significant with the VGG architecture, since in the DenseNet and ResNet architectures the computational need of the pooling operation is proportionally smaller.


\begin{table}[!t]
\caption{This table contains the comparison of different pooling operations on image classification tasks on various datasets. The four columns contain accuracy results for standard maximum pooling (MAXP), Local Importance-based Pooling (LIP),  Wavelet Pooling (WaveP) and continuous pooling (ContP). The accuracy results for Wavelet Pooling are taken from  \cite{williams2018wavelet} and \cite{li2020wavelet}, for Local Importance-based Pooling from \cite{gao2019lip} (except for the cells marked with starts, which contain the accuracy results of our implementations, since no data was available for these datasets and architectures in the literature. The implementation of the LIP operator was taken from \href{https://github.com/sebgao/LIP}{ \url{https://github.com/sebgao/LIP}}). The baseline implementations are also taken from \cite{williams2018wavelet}, \cite{li2020wavelet} and \cite{gao2019lip} and the results in the last column are coming from our implementation. }\label{TableImgClassificationResults}
\label{sample-table-imagenet}
\vspace{-0.5cm}
\vskip -0.15in
\begin{center}
\begin{small}
\begin{sc}
\begin{tabular}{l|c|c|c|c}
\toprule
 & MaxP & LIP &  WaveP & ContP\\
\midrule
\makecell{MNIST \\ LeNet5}  & 95.30   & 96.15 *&   $\mathbf{99.01}$ & 98.87\\
\hline
\makecell{CIFAR-10 \\ AlexNet}  & 71.42 & 77.45* &   74.42 & $\mathbf{81.01}$\\
\hline
\makecell{Imagenet \\ VGG-16}  & 73.37    &  75.12* &   74.40 & $\mathbf{74.65}$\\
\hline
\makecell{Imagenet \\ Densenet-121}  & 74.65    & 76.64  &  75.44  & $\mathbf{76.71}$\\
\hline
\makecell{Imagenet \\ ResNet50}   & 76.15   &  78.19   &  76.71 & $\mathbf{78.23}$\\
\hline
\makecell{Imagenet \\ ResNet101}   & 77.37   &  $\mathbf{79.33}$  &  78.51   & 79.04\\
\bottomrule
\end{tabular}
\end{sc}
\end{small}
\end{center}
\vskip -0.3in
\end{table}

\subsection{Instance segmentation on MS-COCO}

To test our method apart from classification tasks we have also applied it for instance segmentation and object localization on the MS-COCO \cite{lin2014microsoft} dataset.
We have selected the Detectron2 \cite{wu2019detectron2} framework for evaluation and used MASK R-CNN with ResNext-101 backbone \cite{xie2017aggregated}  with feature pyramid network. Training was executed for  270,000 iterations, with 2 images per batch with two variants of the backbone network, one containing maximum the other  continuous pooling with $10$ discretized iterations and  $P_{S_t}$ parameters were all initially set to $0.1$. The average precision results are displayed in Table \ref{TableMaskRCNN}. As it can be seen from the results the test accuracies at the end of both trainings are better using continuous pooling. We also have to note that continuous pooling performed worse in case $AP_{50}$ for bounding box regression at early iterations, but the final accuracy is above the maximum pooling version. We hypothesize that the location sensitivity of continuous pooling helps the network to improve the localization of the objects and the more accurate regression of bounding boxes.

\begin{table}[!t]
\caption{Test accuracies for Mask-RCNN on MS-COCO on Segmentation (Seg) and object detection with bounding boxes (Box) tasks at different iterations (50000, 100000, 150000 and 270000). The columns show mean average precision at $IoU=50:.05:.95$ ($AP$) and average precision at $IoU=0.5$ ($AP_{50}$) with maxim pooling (MP) and its continuous counterpart (CP)  }\label{TableMaskRCNN}
\label{sample-table-coco}
\vspace{-0.5cm}
\vskip -0.15in
\begin{center}
\begin{small}
\begin{sc}
\begin{tabular}{l|c|c|c|c}
\toprule
 & MP-$AP$ & MP-$AP_{50}$ & CP-$AP$ & CP-$AP_{50}$\\
\midrule
Seg(50k)   & 23.87& 44.56 & 25.94 &  44.77\\
Seg(100k)  & 26.86  & 45.55 & 27.52& 48.62 \\
Seg(150k) & 28.66&  51.80 &  30.95 &  54.35\\
Seg(270k)  & 36.47 & 58.07 & 37.12 &  59.11\\
\hline
Box(50k)   & 23.63 & 41.90 & 23.89  & 40.12\\
Box(100k)  & 28.16 & 48.43 & 30.42 &  44.98\\
Box(150k)  & 30.53 & 50.79  & 36.57 & 53.34\\
Box(270k)  & 40.01 & 61.32 & 41.27 &  62.41 \\
\bottomrule
\end{tabular}
\end{sc}
\end{small}
\end{center}
\vskip -0.1in
\vspace{-0.5cm}
\end{table}

\vspace{-0.2cm}
\section{Quantization of the receptive fields}\label{REcFieldQuant}

We have demonstrated in the previous section that the application of continuous pooling can increase network accuracy in classification, object detection and instance segmentation tasks. Unfortunately we have also shown that this increase in accuracy results a growth in computational complexity as well and the number of floating point operations in the network can increase by $20-90\%$ depending on the network architecture.
An increase with other methods such as Wavelet Pooling or LIP can be observed as well, but it is less significant ($25-40\%$)\cite{li2020wavelet} \cite{gao2019lip}.

In this section we would like to demonstrate that the main advantage of our method comes from the optimization of the receptive fields of the neurons not only from the different behaviour of the pooling operator. Because of this we have trained the  DenseNet-121 and ResNet-50 architectures with continuous pooling and once training was finished we have quantized the receptive field in each pooling operation for each convolutional channel (since we were using a different $P_{S_t}$ parameter for each channel) and reverted them back to traditional maximum pooling.
For this we have generated a Dirac-delta input (similarly to Fig. 3) and calculated its activation map after the continuous pooling operation. After this we have thresholded the output activation at $50\%$ of the maximum intensity and from the radius of the resulted sphere we calculated the closest maximum pooling range, which was between 1 and 10.
Using these steps we have reverted back our continuous approach and approximated our continuous pooling operation with maximum pooling. 

To avoid the one-by-one application of maximum pooling with different window size for each channel, we have reordered the channels according to their pooling window size, grouping the same sizes together.
We have implemented the pooling operations for each group according to their window size and reorganized them after pooling to their original order, similarly how it is implemented in \cite{wu2018shift}.
Once this modification of the architecture was done, we have continued training for two epochs.
The accuracy and computational need of this approach compared to the original pooling algorithms can be seen in Table \ref{TableComCost}.
By this we have demonstrated that a significant part of the improvement is coming from the change in the network architecture: a different sized pooling window can be used for each convolutional channel. With this method we could also decrease the computational complexity of the networks resulting only a minor drop in accuracy. Additionally this also means that the receptive size optimization could be combined with other pooling methods (e.g. LIPS), substituting the continuous approach with these algorithms instead of maximum pooling, but the investigation of this was out of the scope of the current paper.

\begin{table}[!t]
\caption{This table contains the accuracy and computational cost of the DenseNet-121 and ResNet-50 architectures on ImageNet, for traditional maximum pooling (MaxP), Local Importance-based Pooling (LIP), Continuous Pooling (ContP) and the qunatized variant of Continuous Pooling (QuantP), where the receptive fields are quantized and reverted back to maximum pooling operations. }\label{TableComCost}
\label{sample-table-quantized}
\vspace{-0.5cm}
\vskip 0.15in
\begin{center}
\begin{small}
\begin{sc}
\begin{tabular}{l|c|c|c|c|}
\toprule
   &MaxP&  LIP &  ContP &  QuantP  \\
\midrule
\makecell{ Densenet-121\\Accuracy\\GFLOP}  & \makecell{ \\74.65\\ 2.88}    & \makecell{ \\76.64\\ 4.13}  &  \makecell{ \\76.71\\ 5.18} & \makecell{ \\76.12\\ 2.95} \\
\hline
\makecell{ResNet-50\\Accuracy\\GFLOP}  & \makecell{ \\76.15\\ 4.12}    &  \makecell{ \\78.19\\ 5.33}  &   \makecell{ \\78.23\\ 6.12} & \makecell{ \\77.24\\ 4.27}\\
\bottomrule
\end{tabular}
\end{sc}
\end{small}
\end{center}
\vskip -0.1in
\vspace{-0.5cm}
\end{table}

\vspace{-0.2cm}
\section{Conclusion}\label{SecConclusion}

In this paper we have demonstrated certain limitations of maximum pooling and introduced a novel approach: continuous time pooling which instead of selection, propagates larger activations in a neighborhood by diffusion in continuous time.
We have demonstrated how our differential equation can be approximated on boolean hardware using the Euler method, since differential equations can not be efficiently implemented on current architectures. Approximation results an iterative approach of maximum pooling which still alleviates the quantization problems caused by maximum pooling and the kernel size can be optimized by traditionally applied gradient based methods.

We compared our method to other pooling operations and have demonstrated on commonly applied datasets and architectures that the continuous time implementation, which also reflects the position of an activation inside a pooling region, increases the overall accuracy of the network. Our method can also be quantized and reverted back to maximum pooling after training, resulting a balance between accuracy and computational need and opens the possibility of combination with other pooling operators.

\vspace{-0.2cm}
\section*{Acknowledgements}
\vspace{-0.2cm}
The support of grants 2018-1.2.1-NKP-00008 Exploring the Mathematical Foundations of Artificial Intelligence and  EFOP-3.6.2-16-2017-00013 are gratefully acknowledged.

\bibliographystyle{IEEEtran}
\bibliography{continuous.bib}

\end{document}